# Towards an optimised evaluation of teachers' discourse: The case of engaging messages

**Samuel Falcon**: [1]University of Las Palmas de Gran Canaria, Department of Education, [2]Instituto Universitario de Análisis y Aplicaciones Textuales (IATEXT), University of Las Palmas de Gran Canaria. samuel.falcon@ulpgc.es, https://orcid.org/0000-0003-3314-1945

**Jaime Leon***: [1]University of Las Palmas de Gran Canaria, Department of Education, [2]Instituto Universitario de Análisis y Aplicaciones Textuales (IATEXT), University of Las Palmas de Gran Canaria. jaime.leon@ulpgc.es, https://orcid.org/0000-0002-9587-4047

**Abstract**

Evaluating teachers' skills is crucial for enhancing education quality and student outcomes. Teacher discourse, significantly influencing student performance, is a key component. However, coding this discourse can be laborious. This study addresses this issue by introducing a new methodology for optimising the assessment of teacher discourse. The research consisted of two studies, both within the framework of engaging messages used by secondary education teachers. The first study involved training two large language models on real-world examples from audio-recorded lessons over two academic years to identify and classify the engaging messages from the lessons' transcripts. This resulted in sensitivities of 84.31% and 91.11%, and specificities of 97.69% and 86.36% in identification and classification, respectively. The second study applied these models to transcripts of audio-recorded lessons from a third academic year to examine the frequency and distribution of message types by educational level and moment of the academic year. Results showed teachers predominantly use messages emphasising engagement benefits, linked to improved outcomes, while one-third highlighted non-engagement disadvantages, associated with increased anxiety. The use of engaging messages declined in Grade 12 and towards the academic year's end. These findings suggest potential interventions to optimise engaging message use, enhancing teaching quality and student outcomes.

Keywords: Teacher Evaluation; Secondary Education; Discourse; Artificial Intelligence.

# 1. Introduction

In recent years, the evaluation of teachers has become a central focus in educational research, driven by the need to enhance teaching quality (Charalambous & Praetorius, 2020; Chen et al., 2021; Harrison et al., 2022; Hopfenbeck, 2024; Looney, 2011). High-quality professional development for teachers can facilitate the learning of best teaching practices, which in turn can lead to higher levels of student performance (Borko et al., 2010; Didion et al., 2020; Gore et al., 2021; Hubers et al., 2022; Schelling & Rubenstein, 2023). For instance, feedback on actual practices has proven effective in enhancing teaching methods and subsequently improving student outcomes (Allen et al., 2011; Gregory et al., 2017), even among students not directly taught by the teachers receiving the feedback (Opper, 2019). Thus, focusing on the evaluation of teaching practices to facilitate professional development is essential, as it can lead to improved teaching methods and ultimately to higher levels of student outcomes.

Despite its acknowledged importance and the pressures from high-stakes accountability systems, most professional development opportunities remain fragmented and insufficient to meet teachers' needs (Borko, 2004; Hsu & Malkin, 2013). The reason for this may be that, although it is known that teaching practices such as cognitive activation, supportive climate, and classroom management, are relevant for enhancing teaching quality and student outcomes (Xie & Derakhshan, 2021), these dimensions may be too abstract or general, which can hinder the implementation of concrete actions to improve teaching quality. In this regard, evidence suggests that targeting more specific factors for intervention, rather than abstract ones, allows teachers to better understand and change their practices (Soderberg et al., 2015).

One aspect of teaching quality that is specific enough to be actionable, does not demand extensive time for improvement (Hardman, 2016; Xie & Derakhshan, 2021), and has the power to shape students' outcomes (Caldarella et al., 2023; Howe & Abedin, 2013; Mercer, 2010) is teacher's discourse. This discourse can be broadly divided into instructional time, which focuses on delivering academic content, and non-instructional time, involving interactions with students not directly related to teaching the curriculum (Dale et al., 2022; Nurmi, 2012; Parsons et al., 2018). While instructional time is crucial, the interactions during non-instructional time, constituting about 20% of teachers' discourse (OECD, 2019), also play a significant role in the overall educational experience, impacting various aspects of students' educational experiences.

Within these interactions, recent studies highlight the importance of the messages that teachers use with their students due to their influence on student outcomes. For example, Putwain et al. (2021) found that messages highlighting the consequences of failure before high-stakes exams led to lower motivation in secondary education students, which then related with poorer performance. Conversely, other studies have found that positive teacher messages, such as those that emphasise the benefits of engaging in school tasks (Santana-Monagas et al., 2023; Santana-Monagas, Núñez, et al., 2022; Santana-Monagas & Núñez, 2022), or those aimed at praising students (Caldarella et al., 2023; Spilt et al., 2016), are associated with improved student behaviour and performance. Given the impact of teachers' messages on student outcomes, it is evident the utility of measuring the actual use of these messages. This measurement could provide teachers with feedback based on their actual practices to facilitate the improvement of their discourse and, in turn, the enhancement of student outcomes (Klentschy, 2005).

To effectively explore the actual use of messages, it would be useful to employ open-ended data collection methods that go beyond traditional Likert-type scales, which primarily focus on students' perceptions rather than teachers' actual practices and do not reflect the real usage of the messages (Spooren et al., 2013; Urdan, 2004). In this context, audio recordings of lessons offer a viable alternative, as they allow to capture everything the teacher says in the classroom. However, this methodology is often limited by the associated time-consuming and labour-intensive coding, which makes it difficult to analyse large amounts of data (Rahman, 2016; Tempelaar et al., 2020).

To address these limitations, the present study leverages recent advancements in the semi-automatic analysis of teacher discourse through automatic audio transcription (Dale et al., 2022; Falcon et al., 2024) and the latest progress in large language models (Demszky et al., 2023) to develop a methodology for optimising the analysis of audio recorded lessons in order to study the teachers' discourse. This approach aims to allow researchers to obtain direct observations of actual message usage, overcoming the traditional limitations of lengthy coding times.

To conduct this study, we have focussed on engaging messages, a type of teacher message that has gained significance in recent years due to its impact on students' motivation, vitality, and performance (Santana-Monagas, Putwain, et al., 2022; Santana-Monagas et al., 2023; Santana-Monagas & Núñez, 2022). By doing so, the study will contribute methodologically by advancing the techniques used for analysing teacher discourse and theoretically by deepening the understanding of how teachers use engaging messages.

**1.1 Teachers' engaging messages**

Teachers' engaging messages are those used to engage students in school tasks (Falcon & Leon, 2023; Santana-Monagas & Núñez, 2022). Some examples of these messages include "*If you study hard enough, you will be able to get into the degree you*

*want*", and "*If you don't do your homework, you will be punished with a lower grade*". These examples show that engaging messages can emphasise either the positive outcomes of engagement or the negative consequences of non-engagement. This characteristic has been extensively explored by the Message Framing Theory (Rothman & Salovey, 1997).

According to this theory, messages can produce different outcomes depending on whether they highlight the benefits of engaging in a specific activity (gain-framed) or the negative consequences of not engaging in the activity (loss-framed; Rothman & Salovey, 1997). Research indicates that gain-framed messages are generally more effective in fostering people engagement (Tversky & Kahneman, 1986). For instance, in the medical field, patients are more likely to opt for surgery when doctors highlight the benefits rather than the potential drawbacks (Rothman et al., 2006). Similarly, in education, there is evidence that loss-framed messages, often used around high-stakes exams, can reduce student motivation and performance (Putwain et al., 2021; Putwain & Best, 2011; Putwain & Symes, 2011). From these results, we can conclude that it would be advantageous for teachers to use gain-framed messages to encourage student engagement in school tasks, rather than relying on loss-framed messages.

Several studies on engaging messages support this conclusion. For example, Santana-Monagas, Putwain, et al. (2022) demonstrated that students' perception of the use of gain-framed messages was associated with higher motivation to learn, subsequently predicting better academic performance. Additionally, Santana-Monagas et al. (2023) indicated that students' perceptions of teachers' use of gain-framed messages enhanced student vitality. Lastly, Santana-Monagas & Núñez (2022) suggest that gain-framed messages are more advantageous in promoting students' psychological need satisfaction, well-being and perseverance. Given this evidence, engaging messages emerge as a promising aspect of teacher discourse to be studied.

**1.2 Measuring teachers' actual use of engaging messages**

The studies discussed in the previous section regarding the impact of engaging messages on students rely on student self-reports as a method of measurement. However, these instruments provide insights into students' perceptions of message use rather than the actual use of such messages by teachers (Urdan, 2004). Therefore, it would be advantageous to examine teachers' actual use of these messages and investigate their relationship with student outcomes (Klentschy, 2005; Soderberg et al., 2015). In this regard, using audio recordings of lessons could be a particularly effective approach, as this method allow us to capture everything the teacher says (Dale et al., 2022). Thus, this method provides us with a means of capturing accurate observations of teachers' use of engaging messages.

Audio-recorded lessons can be analysed through direct listening or, more commonly, by examining their transcripts, a process known as transcript-based lesson analysis (Arani, 2017). This method has become an important tool in educational research, enabling more efficient analysis of discourse compared to traditional methods of direct observation (Demszky & Hill, 2022; Winarti et al., 2021). By allowing to provide insights into actual teaching practices, transcript-based lesson analysis contributes to improving teaching quality and facilitates significant professional development for teachers (Chen et al., 2014; Richards & Farrell, 2011; Soderberg et al., 2015). Therefore, the examination of transcripts of audio recordings emerges as the most appropriate method for exploring teachers' actual use of engaging messages.

The first step in this analysis involves transcribing the teacher's speech into text. Traditionally done manually, this process can be quite time-consuming, often requiring two to four hours to transcribe a single 45-minute class (Dale et al., 2022). Consequently, most researchers now use automatic transcription systems. For example, Demszky & Hill (2022) automatically transcribed 1660 elementary mathematics lessons, and Dale et al. (2022) did the same for 127 English Language Arts lessons, allowing for a and more efficient analysis of teachers' discourse. Accordingly, this methodology was adopted for transcribing the audio-recorded lessons in our study.

Despite the efficiency of automatic transcription, manual coding of transcripts is still prevalent and often requires several hours of analysis per lesson (Dale et al., 2022). To address this problem, researchers typically rely on training multiple assistants for coding, distributing the workload and streamlining the process (Rahayu et al., 2020). However, this method depends on the availability of assistants and can be lengthy, with decision fatigue potentially compromising analysis quality (Vohs et al., 2005). This highlights the need to further optimise transcript analysis to handle larger datasets more efficiently.

Seeking more efficient alternatives, the recent study conducted by Demszky et al. (2023) highlights the utility of large language models (LLMs) in facilitating the coding of transcripts by fine-tuning these models with real-world examples. Fine-tuning refers to the process of re-training a model using specific data to improve its performance for a particular task. In our context, this would mean training LLMs using examples of actual engaging messages for teaching it to identify similar messages in the transcripts and classify them as gain- or loss-framed messages. Therefore, the application of fine-tuned LLMs could simplify and optimise the identification and categorisation of engaging messages within classroom discourse transcripts.

**1.3 Purpose of the research**

By combining the automatic transcription of audio-recorded lessons with the analysis of transcripts through fine-tuned LLMs, we aim to offer a methodology that allows a more optimised study of teachers' discourse and avoid the traditional limitations of manually coding this type of data.

To achieve our aims, we present two studies. Study 1 utilises a database with real examples of engaging message compiled over two academic years to fulfil three objectives: firstly, to fine-tune an LLM for identifying engaging messages within lesson transcripts; secondly, to fine-tune a second LLM to classify the identified engaging messages as gain- or loss-framed messages; and lastly, to analyse the performance of these models to assess their feasibility for analysing transcripts for the identification and classification of engaging messages. Therefore, this study aims to contribute to the study of teacher discourse by establishing a robust methodology that facilitates the analysis of audio-recorded lesson transcripts.

Study 2 aims to use the fine-tuned models from Study 1 to extract observations of the engaging messages used by teachers in a different dataset, from a third academic year. This study seeks to demonstrate the utility of the fine-tuned LLMs in coding actual observations of engaging messages from a large-scale dataset. Additionally, we aim to deepen our understanding of their use by studying the frequence of the messages, examining how their use varies across educational levels, and throughout the academic year. This research aims to contribute to the study of engaging messages, providing insights on their actual use and establishing the basis for future inferential studies that use actual observations of messages rather than student perceptions of their use. Additionally, we hope to establish guidelines for other studies interested in examining teacher discourse using this approach, ultimately contributing to more effective teaching practices and improved student outcomes.

## 2. Study 1

### 2.1 Methods

#### *2.1.1 Participants*

Study 1 involved a cohort of 74 secondary school teachers (females=41, mean age=45.59, SD=8.74) among two academic years. These participants were stationed at various public secondary schools situated in five distinct regions throughout Spain, covering both urban and rural areas. Most of the educators participated in the study with several groups they were instructing, resulting in 126 participant groups. The breakdown of groups by grade level included 41 in Grade 9, 30 in Grade 10, 18 in Grade 11, and 37 in Grade 12.

#### *2.1.2 Procedure*

At the beginning of each academic year, principal heads of schools were contacted for the purpose of enlisting participant teachers. The objectives of the study were clearly conveyed to the interested teachers, with emphasis on the voluntary nature of their participation and the guarantee of confidentiality. This was further affirmed by securing their agreement through an 'informed consent' document. The study remained compliant with the data protection laws, regulations, and standards at both the national and European levels.

The academic schedule in Spain is divided into three trimesters. We asked teachers to record at least eight of their lessons in each trimester. We provided them with compact recording devices (8 cm) that were easy to use and could be conveniently attached to their clothes near their mouth. These devices were configured to capture nearby sounds, ensuring that when placed near the teacher's mouth, only their voice was recorded. Teachers were given demonstrations on the usage of the recording equipment and were instructed to forward the audio files to us at the end of each term. The final number of audio files received over the two academic years was 2,516. We then transcribed the audios using Whisper, an open-source model for automatic transcription that has already proven useful in transcribing lesson recordings (Bain et al., 2023; Rao, 2023).

*2.1.3 Creation of the dataset*

To create the engaging messages dataset[1] for fine-tuning the LLMs, we used a combination of automatic transcription and keyword-based filtering described in previous studies (Falcon, Admiraal, et al., 2023; Falcon et al., 2024). The first step of this approach was to assess the presence of engaging messages from teachers in the lessons' transcripts. The second step was to manually classify the messages identified in the above step as gain- or loss-framed messages. See Supplemental Material Appendix for a full explanation of the procedure. The final number of messages detected over the two academic years was 856.

*2.1.4 Fine-tuning of the models*

Of all the LLMs available at the time of this study, we decided to choose the Ada version of GPT-3 for fine-tuning both the identification and classification models[2] due to its suitability for binary classification (Brown et al., 2020; OpenAI, 2023). The identification model analyses the complete transcription of lessons and labels each sentence as an engaging message (1) or not an engaging message (0) based on the examples with which it was fine-tuned. The classification model then analyses the messages identified by the first model and classifies them as either gain-framed (1) or loss-framed (0) according to their similarity to the examples with which it was fine-tuned.

---

[1] The dataset is available at: https://osf.io/m5evw/?view_only=10881b23f70e45d2834335f8805d0aec
[2] The programs are available at: https://osf.io/vtpwn/?view_only=7704c0a143014d8e9867a44a5b530d79

For the fine-tuning process of the identification model, we created two Python programs (See Supplemental Material, Figure S1). The first program fine-tuned the model using two datasets: one with all transcript sentences from the two academic years and another with the 856 engaging messages described in the previous section. First, it created a dataset of positive (engaging messages) and negative (non-engaging sentences) examples from the two original datasets. We established a 5/95 ratio for our example dataset, resulting in 856 engaging messages and 16,264 non-engaging sentences. The dataset was split into training, validation, and testing sub datasets at an 80/10/10 ratio. The model was then fine-tuned using Azure OpenAI Studio and employing OpenAI's recommendation for the hyperparameters (Microsoft, 2023; OpenAI, 2023). The second program tested the fine-tuned model using the testing subset. For this, the actual label of each sentence (i.e., 0 or 1) was masked, and the model was asked to analyse each sentence to determine if it was an engaging message or not. Finally, the model's results were compared with the correct classifications, and a reliability analysis was conducted.

For the fine-tuning process of the classification model, the steps were very similar but with some changes, which led us to create two other programs (See Supplemental Material, Figure S2). The first program only used the 856 engaging messages dataset from the two academic years. The first step performed by the program was to split the 856 messages dataset into gain-framed messages (479) and loss-framed messages (377). The dataset was then split at an 80/10/10 ratio. Again, we used the Ada model for fine-tuning and the same hyperparameters. The second program facilitated testing using the fine-tuned model and the testing sub dataset. Finally, we compared the results with the correct labels, and we performed a reliability analysis.

*2.1.5 Data analysis*

To assess the performance of the fine-tuned models, we employed several metrics commonly used for evaluating classification tasks in AI models: sensitivity (or recall), specificity, precision, and F1 score (Goutte & Gaussier, 2005; Shahriari et al., 2023). Sensitivity and specificity, originally derived from clinical tests designed to confirm or refute the presence of a disease, help us evaluate a model's accuracy in identifying and classifying engaging messages. Additionally, precision and the F1 score provide further insight into the balance between the models' predictive accuracy and their ability to minimise false positives and false negatives. See the Supplemental Material Appendix for a detailed explanation of these parameters.

In the identification model, sensitivity (or recall) measures the proportion of actual engaging messages correctly identified by the model, indicating its ability to detect true positives (sentences that are engaging messages). High sensitivity means the model is effective at identifying engaging messages. Specificity, on the other hand, measures the proportion of non-engaging sentences correctly classified, reflecting the model's accuracy in identifying true negatives (sentences that are not engaging messages). High specificity ensures the model is reliable in excluding non-engaging sentences. Precision refers to the proportion of sentences predicted to be engaging messages that are actually engaging. Therefore, high precision indicates that the model makes few false positive errors, ensuring that when the model predicts a sentence as engaging, it is more likely to be correct. Lastly, the F1 score combines precision and sensitivity (recall) into a single metric, being useful when there is an imbalance between engaging messages and non-engaging sentences. A high F1 score indicates that the model achieves a good balance between correctly identifying engaging messages and avoiding false positives.

For the classification model, gain-framed messages were labelled as 1s, and loss-framed messages were labelled as 0s in the dataset. This means that the model was fine-tuned for the classification of gain-framed messages. Therefore, in this case sensitivity can be understood as the proportion of gain-framed messages correctly classified by the model, while specificity is the proportion of loss-framed messages correctly classified. Precision in the classification model represents the proportion of messages predicted as gain-framed that are actually gain-framed. The F1 score, as in the identification model, provides a comprehensive measure of the classification model's performance by balancing precision and sensitivity. A high F1 score demonstrates the model's ability to consistently and accurately classify gain-framed messages while avoiding false positives.

## 2.2 Results

The fine-tuning process for the identification of messages in the transcripts task took 2.5 hours and the resulting model demonstrated strong performance. When evaluated on the testing dataset comprising 80 engaging messages and 1,576 non-engaging sentences, the model achieved a sensitivity of 84.31%, a specificity of 97.69%, a precision of 64.4% and a F1 Score of .73. Specifically, of the 80 engaging messages in the testing sub dataset, the model correctly classified 67 (true positives). The sensitivity of 84.31% reflected a low number of false negatives, with only 13 of 80 engaging messages mislabelled as non-engaging by the model. Table 1 shows some examples of the 13 cases in which the model was not able to correctly identify the messages. The high specificity also indicated the model's proficiency in correctly labelling non-engaging sentences, with 1,539 of 1,576 accurately classified as non-engaging sentences (true negatives). The F1 score of .73 suggests that the model effectively identifies engaging messages while maintaining a good balance between avoiding false positives and minimising false negatives. Overall, the fine-tuned model exhibited good results in distinguishing engaging messages from other discourse content.

**Table 1. Examples of false negatives in the identification phase**

| Sentence analysed by the fine-tuned LLM | Engaging message identified by the human coders |
|---|---|
| this guys, what you are studying now is | this guys, what you are studying now is basic for the coming months. So, whoever doesn't know about this, please start studying, because if you don't know how to do this, then it gets complicated, it gets complicated, and you fail. |
| more work nights or not | with a little more work, you'll raise that grade a lot |
| it is normal if you find it difficult and complicated the first time | it is normal if you find it difficult and complicated the first time. I have done it a lot of times and that is why it is easier for me, but knowing how to solve problems is a quality that will be useful to you forever, I think it is worth an effort. |

*Note*. These sentences are translated from Spanish.

From Table 1, we can observe that the missed engaging messages were incomplete or inaccurate sentences, a consequence likely attributable to the quality of the original audio records. In the case of the false positives, they were mainly long, well-structured sentences featuring words associated with engaging messages, but which were not genuinely conveying an intention to engage the students. For instance, the following false positive contained elements that could be present in an engaging message, such as "professional opportunities" or "choosing a career": "*I imagine that you have been provided information about the baccalaureate and professional opportunities, the criteria for choosing one career or another*".

The process of fine-tuning for the classification of messages in the frame categories lasted 25 minutes. The testing sub dataset consisted of 45 gain-framed and 44 loss-framed messages. The fine-tuned model achieved a sensitivity of 91.11% by correctly identifying 41 of the 45 gain-framed messages. It also demonstrated a specificity of 86.36% by accurately labelling 38 of the 44 messages as loss-framed messages. The high sensitivity indicated a low number of false negatives, with only 4 gain-framed messages incorrectly classified. The specificity also reflected a small number of false positives, as just 6 loss-framed messages were mislabelled. In addition to sensitivity and specificity, the model achieved a precision of 87.2% and an F1 score of .89. highlighting the overall strong performance of the model.

**2.3 Discussion**

Study 1 demonstrates the potential of using fine-tuned LLMs in combination with automatic transcription for optimising the analysis of teachers' discourse. Our findings indicate that the fine-tuned model on a dataset of engaging messages and non-engaging sentences can effectively identify these messages within lesson transcripts with high sensitivity and specificity. The model achieved 84.31% sensitivity and

97.69% specificity on the testing sub dataset, demonstrating strong performance in detecting engaging messages and avoiding other types of sentences. The performance of the fine-tuned classification model was also promising. Sensitivity and specificity exceeded 85% in distinguishing between gain-framed and loss-framed messages, indicating that the model can reliably categorise the frame once messages are identified. Both of these results are promising, suggesting that fine-tuned LLMs could be used to optimise the coding process, saving time and resources for researchers interested in analysing teacher discourse.

These findings align with recent perspectives indicating the promise of leveraging advancements in LLMs to efficiently analyse transcripts and identify key constructs (Demszky et al., 2023), facilitating the linking of educational theories to real-world teacher discourse and enhancing the practical application of theoretical knowledge.

*2.3.1. Limitations and future directions*

Although the model demonstrated satisfactory performance in identifying engaging messages, training with higher quality transcriptions could potentially enhance its accuracy further. The impact of transcript quality on analysis is well-established; higher quality transcripts invariably lead to better outcomes (Janah et al., 2019). However, obtaining high quality recordings can be complicated when gathering data from naturalistic environments. Therefore, future studies should take this into account and make an effort to capture participants' voices clearly.

The fine-tuned LLM identified some false positives of sentences that were not engaging messages as actual messages. These instances included terminology associated with engaging messages (e.g., "career opportunities") but lacked the genuine intent of engaging students. To address this, increasing the examples dataset might help the model better differentiate between engaging messages and other types of sentences. Beyond collecting more data, employing techniques such as data augmentation or generative methods (Argyle et al., 2023) would likely improve the model's already promising performance.

Expanding the analysis to other factors of teachers' discourse, both instructional and non-instructional, is a natural next step that can demonstrate wider feasibility of the method. For instance, fine-tuned LLMs offer an efficient approach to systematically obtaining actual observations of empathetic messages (Eisenberg et al., 2010; Meyers et al., 2019; Patel, 2023), motivational messages (Alqassab & León, 2024; Putwain et al., 2021), and praise (Jenkins et al., 2015; Lipnevich et al., 2023), among other verbal factors of teachers' discourse (Metsäpelto et al., 2022). Therefore, fine-tuned LLMs hold significant promise for advancing the study of teacher-student interactions.

Lastly, advances in LLMs will offer powerful tools to gain deeper insights. By leveraging state-of-the-art models (i.e., GPT-4o, Gemini, Llama 3.2., etc.), future research can refine this methodology to examine messages or other factors in discourse more effectively. Efficiently analysing large volumes of discourse data could enable scalable interventions using personalised feedback, as demonstrated in past work (Milkman et al., 2021). This approach could be particularly beneficial, as receiving immediate feedback on one's practices has been shown to facilitate improvement (Laranjo et al., 2021). Therefore, the methodology developed in this study can support the development of both pre-service teachers during their training and in-service teachers, enhancing their use of effective discourse and subsequently improving students' outcomes (Borko, 2004; Borko et al., 2010; Gore et al., 2021).

## 3. Study 2

### 3.1 Methods

*3.1.1 Participants*

Study 2 involved a cohort of 19 secondary school teachers (females=9, mean age=46.85, SD=9.58) in one academic year. These participants were also stationed at various public secondary schools situated in four distinct regions throughout Spain, covering both urban and rural areas. These teachers contributed data from 30 different groups they were teaching, which included students from Grades 10 to 12. Specifically, the groups comprised 7 in Grade 10, 9 in Grade 11, and 14 in Grade 12.

*3.1.2 Procedure*

Similarly to study 1, we approached school principals at the beginning of the academic year to enlist participants. Teachers were informed about the study's aims, the voluntary nature of participation, and confidentiality assurances. The same 'informed consent' form was obtained from all participants, ensuring compliance with national and European data protection laws. Teachers were asked to record at least eight lessons per trimester with the same compact recording devices. Teachers submitted the audio files at the end of each trimester, which allowed us to collect 719 audio files over the academic year. These audios were also automatically transcribed using the Whisper model.

*3.1.3 Data analysis*

Each transcription was processed using the fine-tuned model developed in Study 1 to identify engaging messages. These detected engaging messages were subsequently classified as gain- or loss-framed using the second fine-tuned model from Study 1. With the obtained results, we conducted three descriptive data analyses: overall results, results across educational levels, and results throughout the trimesters. For the first and third cases, we calculated the percentage of messages used from each category based on the total message count. For the second analysis, since the number of groups at each educational level differed, we first calculated the following ratio for each type of message: message count divided by the number of groups. Based on this ratio, the percentage of each category of message was then calculated for each educational level.

## 3.2 Results

### 3.2.1 Overall results

After analysing the transcripts with the first fine-tuned LLM to identify engaging messages, we successfully reduced the total number of sentences across the 719 audio file transcripts from an initial 303,418 sentences to just 1,828 sentences. As observed in Study 1, this identification model was not perfect; therefore, these 1,828 sentences contained a mix of engaging messages and false positives (sentences that were not engaging messages). To address this, the two research assistants who performed the manual identification of messages in Study 1 to create the training dataset were tasked with reviewing the 1,828 sentences to eliminate the false positives. This process resulted in an inter-coder agreement of 94.66%. Disagreements were resolved through consensus discussions with the lead researcher, leading to the identification of 202 engaging messages.

Subsequently, these identified messages were classified into gain-framed and loss-framed categories using the second fine-tuned LLM. The classification revealed that 67.33% of the messages were gain-framed, while 32.67% were loss-framed. Notably, both the identification and classification analyses were completed within seconds. Both analyses were carried out within seconds. In Table 2, we show examples of actual engaging messages obtained after the analyses.

**Table 2. Examples of actual engaging messages**

| Frame | Example |
|---|---|
| Gain | "But we are going to try to be alert because what we learn here is time you don't spend at home." |
| | "Just do the summary, and with that summary you are studying and so the 25 questions of the exam will be easier." |
| | "I gave you a general sheet, which contains everything. I think there are a hundred or two hundred exercises there. If you do them all you will learn the subject perfectly." |
| | "Do the notes at home, if you have that opportunity, it can help you to finish off and get, I don't know, a super grade, okay? Do it and that way you can refresh a little bit." |
| Loss | "It is up to you whether you want to lose a point in the university entrance exam for not studying the formulation or not." |
| | "Those of you who are having difficulties, the time to get it is now, otherwise you won't get it with me. You will have to do it on your own, so be careful." |
| | "If you don't pass contemporary world history, you won't get your diploma." |
| | "I hope you will understand that if you don't submit anything to me, even if I want to, I can't pass you." |

*Note*. These sentences are translated from Spanish.

*3.2.2 Results across educational levels*

The total number of messages across educational levels was distributed as follows: Grade 10 = 66, Grade 11 = 75, and Grade 12 = 61. Due to the varying number of groups in each grade, we calculated the ratio and percentage for each level, as detailed in Section 3.1.3. The differences between these levels are illustrated in Figure 1.

The data shows distinct trends for each type of engaging message. Gain-framed messages are more prevalent overall, peaking in Grade 10, whereas loss-framed messages are less common but show a slight increase until Grade 11 before declining. Both types of messages decrease in frequency by Grade 12.

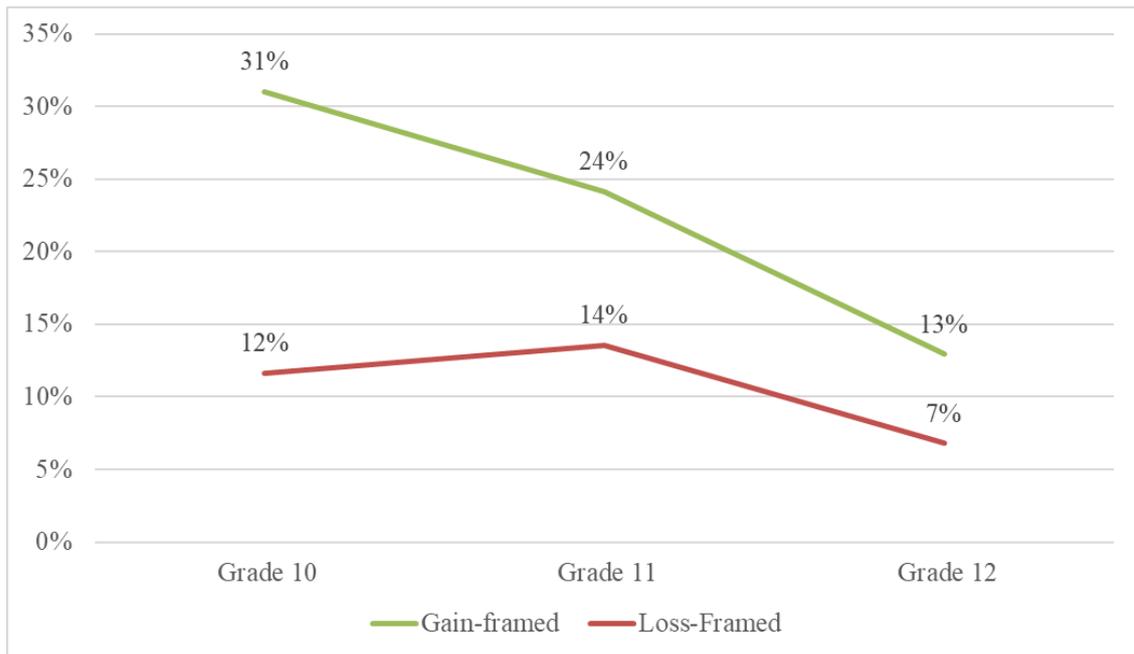

**Figure 1. Percentages of engaging messages found across educational levels.**

### *3.2.3 Results throughout the academic year*

The total number of messages throughout the trimesters was as follows: first trimester = 56, second trimester = 52, and third trimester = 31. Differences between trimesters are illustrated in Figure 2.

The general trend in the data shows a decrease in the use of both types of messages as the academic year progresses. The decline is more pronounced in loss-framed messages, particularly from T1 to T2. However, gain-framed messages, while more prevalent than loss-framed ones throughout all three time periods, also show a consistent downward trend.

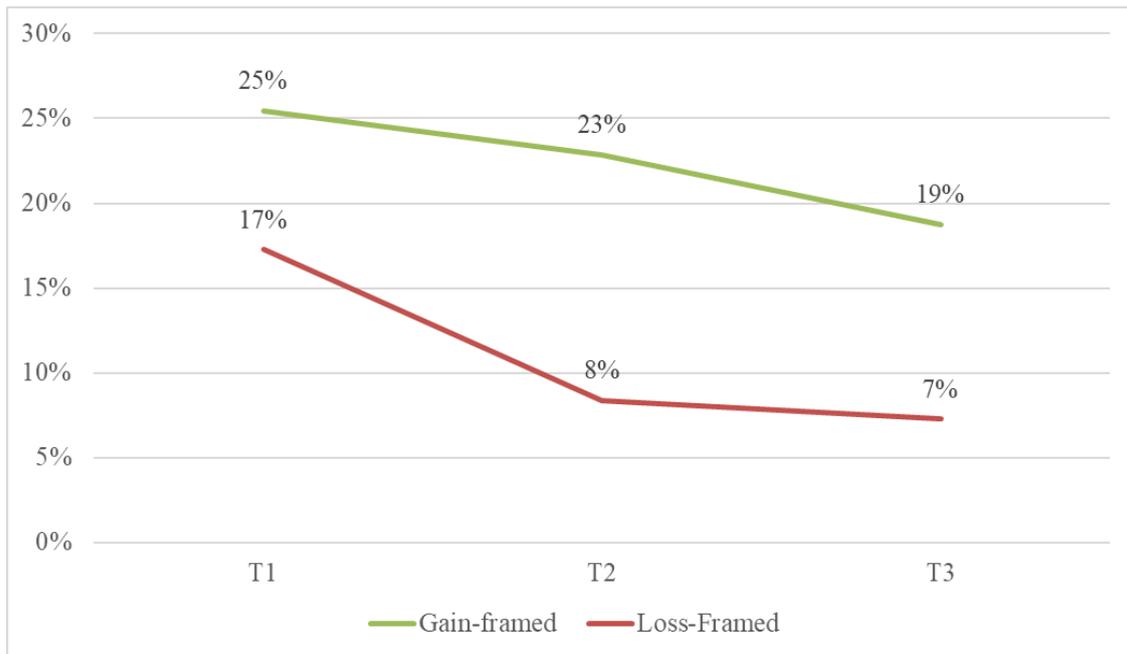

**Figure 2. Percentages of engaging messages found throughout the trimesters.**

### 3.3 Discussion

In Study 2, we used the fine-tuned LLMs from Study 1 to analyse the lessons' transcripts collected throughout the academic year. By using these models, we could reduce the volume of information to be analysed in comparison to other methods (Dale et al., 2022; Falcon, Alonso, et al., 2023; Falcon et al., 2024), enabling us to efficiently identify and extract observations of the actual engaging messages used by teachers. This approach also allowed us to conduct a descriptive analysis of their general usage patterns, and the usage differentiating by educational level and time of the academic year. The overall results indicated that teachers generally use more gain-framed messages than loss-framed ones. This is encouraging, as gain-framed messages are linked to improved motivation, vitality, and student performance (Santana-Monagas et al., 2023; Santana-Monagas, Putwain, et al., 2022; Santana-Monagas & Núñez, 2022). These findings suggest that teachers are employing effective strategies to foster better student engagement, motivation and performance.

However, it is notable that one-third of the messages used were loss-framed, which are associated with increased anxiety and poorer outcomes (Jang & Feng, 2018; Putwain & Remedios, 2014). Despite this, the presence of loss-framed messages is not entirely negative, as recent studies indicate that if students perceive such messages as challenges, they can also be engaging (Putwain et al., 2023). Additionally, while gain-framed messages excel in low-risk scenarios, loss-framed ones might be more effective in higher-risk situations (Jang & Feng, 2018). However, it is important to note that these previous results are based only on students' perceptions of teachers' message use. Therefore, further research could combine the method used in this study for obtaining observations of actual engaging messages with students' self-reports to validate and enhance our understanding of how these messages influence student outcomes.

Regarding educational levels, the findings show that Grade 10 is where the highest prevalence of message usage was found. In Spain, this level marks the end of compulsory education, and students can choose to continue in preparatory levels for university, vocational education, or leave the education system. At this level, teachers used the most gain-framed messages, perhaps because it is a significant transitional stage, and teachers might feel the need to emphasise the benefits of studying to encourage students to continue their education (Jimerson & Haddock, 2015). This might also explain why the percentage of gain-framed messages remains high in Grade 11, possibly to prevent student dropout.

Interestingly, in Grade 11, the level of loss-framed messages was the highest. This increase could be associated with the stress and high stakes of this transitional period, as teachers may resort to emphasising the consequences of not engaging with their studies to motivate students under pressure. Teachers might believe that loss-framed messages can enhance motivation by increasing the perceived threat and urgency of the situation (Janke et al., 2011), or that such messages are more persuasive when decisions involve uncertainty or threat, as in high-stakes educational settings like Grade 11 (Gerend & Sias, 2009).

In Grade 12, there is a surprising decrease in all engaging messages, despite this year in Spain being dedicated to preparing students for university entrance exams, where more gain-framed messages would be particularly beneficial. One possible explanation for this result is that teachers in this level are overwhelmed trying to complete the syllabus to ensure students are well-prepared for these critical exams (Manuel et al., 2018; O'connor & Clarke, 1990), thereby dedicating more time to instruction rather than non-instructional aspects.

By trimester, there is a progressive decline in the use of engaging messages as the school year advances. As before, a possible explanation is that teachers become increasingly burdened with the need to complete the syllabus by the end of the year (Leong & Chick, 2011; Serrano et al., 2013; von der Embse & Mankin, 2021), dedicating more time to instructional content rather than non-instructional aspects.

The insights gained from these observations contribute to a better understanding of teachers' use of engaging messages and can inform future studies aimed at improving their use. Specifically, these findings suggest the need for interventions that consider the educational level and the timing within the school year to identify where more emphasis is required (e.g., in Grade 12 or the third trimester). This targeted approach can help enhance the use of engaging messages, ensuring they are used optimally to support student motivation and performance throughout the academic year.

### *3.3.1. Limitations and future directions*

Regarding the methodology used to obtain the observations, the models used were not perfect, as the performance reported in Study 1 did not achieve 100% sensitivity and specificity. Consequently, some false positives occurred, and some engaging messages were likely missed during the transcript analysis process. However, given the substantial savings in time and resources provided by these models, we believe the trade-off is justified. We encourage future studies to replicate our methodology and conduct further comparisons with other types of messages to determine if they find similar benefits.

Another limitation is that these findings are pertinent to Spanish teachers and should be approached with caution. Cultural variances in teacher motivation and engagement strategies have been noted in prior research (Hagger et al., 2007; Luoto et al., 2023), so the use of these messages may vary depending on the country or culture. To discern potential differences in the use of engaging messages, a broader cross-cultural study encompassing teachers from various nations is essential.

Furthermore, as previously discussed, these advancements could lead to the development of interventions that provide teachers with instant feedback on their use of engaging messages. In this regard, the results of Study 2 can further refine the way these interventions are delivered. For instance, interventions could focus on improving teachers' use of gain-framed messages, particularly in Grade 12, where increased usage could yield significant benefits. Additionally, special emphasis could be placed on encouraging teachers to continue using engaging messages towards the end of the academic year, as the benefits are worthwhile.

## 4. General conclusions

Study 1 demonstrated the potential of fine-tuned LLMs in optimising the identification of engaging messages within lesson transcripts, achieving 84.31% sensitivity and 97.69% specificity, and the classification of those messages in their categories, achieving 91.11% sensitivity and 86.36% specificity. Study 2 applied the fine-tuned LLMs of Study 1 to transcripts of audio recordings from an academic year, revealing that teachers predominantly use gain-framed messages, which are linked to improved student motivation and performance. In conclusion, this research highlights the potential of fine-tuned LLMs for efficient discourse analysis, providing precise information about teachers' actual use of engaging messages. By applying this methodology to further study teachers' use of engaging messages and other aspects of teacher discourse, it will be possible to develop interventions that offer instant feedback on teachers' actual practices and contribute to their professional development. This approach will enhance teaching quality and, consequently, improve student outcomes (Borko et al., 2010; Didion et al., 2020; Gore et al., 2021; Hubers et al., 2022).

# Towards an optimised evaluation of teachers' discourse: The case of engaging messages

SUPPLEMENTAL MATERIAL

**Creation of the dataset used for fine tuning the LLMs**

To create the fine-tuning datasets, the messages in the lesson transcripts were first identified and then classified. In the original procedure of identifying engaging messages, the audio recordings from the lessons of the two academic years were transcribed using the Whisper model (Rao, 2023). This transcription generated around 600 lines of text per lesson, each representing one or several sentences depending on the pauses between them. Then, a pre-established set of keywords was applied to filter through the transcripts using a Python-based program. This list of keywords, developed in earlier studies (Falcon et al., 2023, 2024), comprised terms commonly found in or related to engaging messages (i.e., study, pass, future, etc.). The filtered transcripts, consisting of a concentration of engaging teacher messages mixed with false positives, served as the basis for manually identifying such messages.

A pair of research assistants were trained to identify teacher engaging messages in the filtered transcripts and discard other types of sentences. The established criteria for selecting these messages included: 1. the message intends to engage students in school tasks, 2. the message has a frame, either of gain or loss, and 3. the message has inherent significance, which could encompass one or multiple sentences. The process of identifying engaging messages yielded a good inter-coder agreement rate of 98.71% (O'Connor et al., 2017).

Upon being identified, research assistants corrected the messages for spelling mistakes or if they were cut by the automatic transcription. After this, the assistants systematically categorised these messages based in the two previously defined categories: gain- and loss-framed messages. The categorisations also demonstrated good reliability, with agreement levels above 95% for both categories.

Following the classification, the count of messages in each category was 479 for gain-framed messages, and 377 for loss-framed messages. Examples of each category are presented in Table S1.

**Table S1. Examples of engaging messages**

| Frame | Example |
|---|---|
| Gain | "The more exercises you have solved, the more likely you are to get things right." |
| | "It is clear that if you study, you are more likely to have a decent job." |
| | "Practice and practice and practice to progress." |
| | "Guys do it right and try hard and you will be proud of yourselves after a job well done." |
| Loss | "If you don't pay attention in class, you will be punished without recess." |
| | "If you don't do your homework, you will disappoint me and your parents." |
| | "If you don't work hard, you'll have to make do with studying less sought-after degrees." |
| | "If you don't study, you will miss out on the beauty of this subject." |

*Note*. These sentences are translated from Spanish.

**Fine-tuning of the LLM for message identification**

The first program was tasked with fine-tuning the model for the classification of sentences as engaging messages or non-engaging messages, utilising two datasets for this purpose. The first dataset comprised the 856 sentences classified as engaging messages by the coders as originally found in the filtered transcripts. The second dataset encompassed all sentences from the transcripts of the two academic years, except the 856 identified engaging messages.

In the first step of the program, we determined the ratio of engaging messages and non-engaging sentences for the creation of the example dataset. To illustrate, a 50/50 ratio would make the program to select the 856 engaging messages, along with a random assortment of 856 sentences from the all-encompassing (excluding the 856 identified engaging messages) dataset derived from the two years' transcripts dataset. Following preliminary tests involving various proportions, we observed an enhancement in the model's reliability when the example dataset mirrored the real data, as previous studies have also found (Karp et al., 2012). Consequently, we opted for a 5/95 proportion in our scenario. This implied that the example dataset incorporated 856 engaging messages and 16 264 non-engaging sentences. This dataset was saved in a CSV document with two columns, the first containing the sentence and the second containing a '1', if it was an engaging message, or a '0', if not.

In the second step, we decided the ratios to split the example dataset to form the fine-tuning, validation, and testing sub datasets. The fine-tuning sub dataset consists of examples that the model utilises to learn the designated task – in this case, labelling sentences with a '1' if they are engaging messages, and '0' if otherwise. The validation sub dataset serves to assess the model using data unseen during the training phase in order to adjust the hyperparameters of the model. Lastly, the testing sub dataset enables us to assess the model's performance with additional unseen data. We opted for the 80/10/10 split to divide the example dataset into training, validation, and testing sub datasets, as this is the most common method of segregating datasets for training models in this type of task (Riskiyadi, 2024; Supri et al., 2023; Vrigazova, 2021).

The final stage of the first program involved fine-tuning the model using Azure OpenAI Studio (Microsoft, 2023). In this last step, we selected the model to fine-tune from the various available versions of GPT-3 and determine the tuning hyperparameters. For these two tasks, we relied on OpenAI's recommendations for GPT-3 fine-tuning (OpenAI, 2023) and some preliminary tests. Consequently, we chose to fine-tune the Ada model, which is not only the fastest and most cost-effective but also ideal for classification tasks (OpenAI, 2023). Concerning the hyperparameters, we opted to use a batch size of 16, a learning rate multiplier of 0.1, and a number of epochs set at 20, while also enabling the "compute classification metrics" hyperparameter, which is recommended by OpenAI for classification tasks. After this, the program converted the datasets to the format required by the service (JSONL) and the model was fine-tuned and validated using the corresponding sub datasets to end up obtaining the final fine-tuned model.

The second Python program allowed us to use the testing sub dataset with the newly fine-tuned model while being able to modify the completion parameters. Among these parameters, the two most important for our case are temperature and TopP, which control the level of randomness or creativity that the user wants the model to respond to. In our scenario, as we seek replicable and non-creative responses, we set the temperature to 0 and TopP to 1 (Author et al., 2023). After this, the testing sub dataset is submitted to the fine-tuned model, allowing it to classify the sentences and compare the results with the one previously classified by the human coders. Figure 1 shows a summary of all the process.

**Figure S1. The two Python programs developed for the fine tuning of the model in the identification of messages step.**

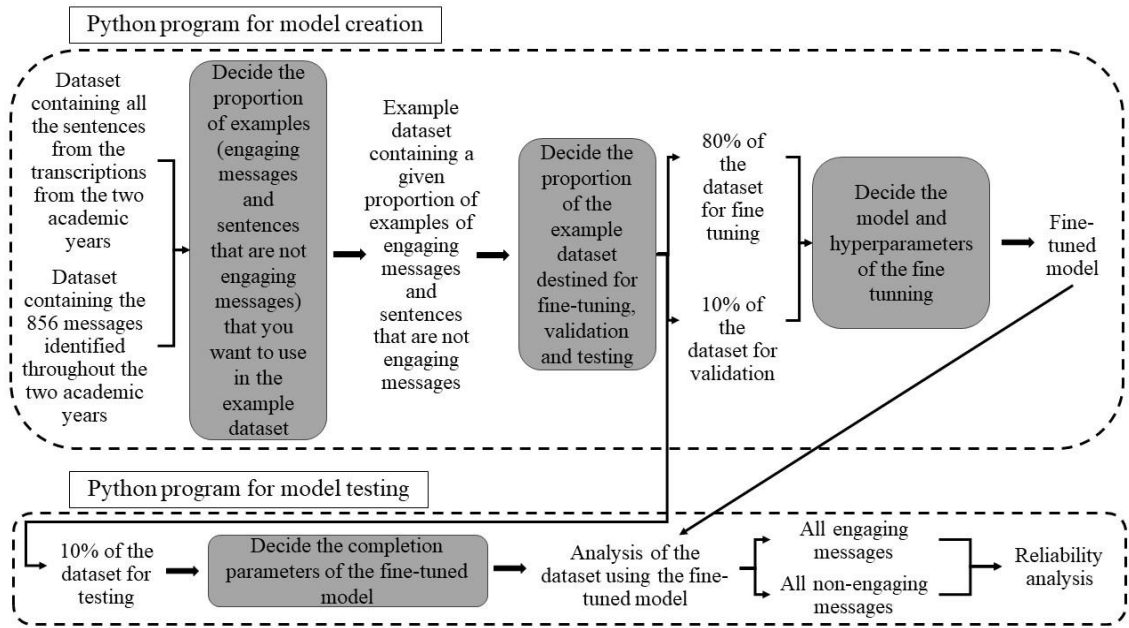

**Fine-tuning of the LLM for message classification**

The first program was also tasked with fine-tuning the model but only utilised the 856 corrected messages pre-classified by human coders as its input. Since the frame is a binary dimension with two categories, only one dataset of examples was created. This dataset had the same structure as the CSV generated during the message identification phase, with one column containing the message and another indicating "1" for gain-framed or "0" for loss-framed.

Following this, similar to the prior identification step, we determined the ratios for creating the fine-tuning, validation, and testing sub-datasets. We chose the Ada model for fine-tuning as the task remained consistent—binary classification of text. Using the same hyperparameters, the datasets were converted to JSONL format, and the model was fine-tuned.

The second Python program served a similar function as in the message identification phase, enabling us to employ the testing sub-dataset with the fine-tuned model and adjust the completion parameters. Given that the objective was to obtain replicable and non-creative responses, we configured the temperature to 0 and TopP to 1. Subsequently, the testing sub-dataset was dispatched to the fine-tuned model. This facilitated the classification of messages into both categories of the frame dimension, allowing for a comparison with the results previously categorised by human coders. Figure 2 shows a summary of all the process.

**Figure S2. The two Python programs developed for the fine tuning of the model in the classification of messages step.**

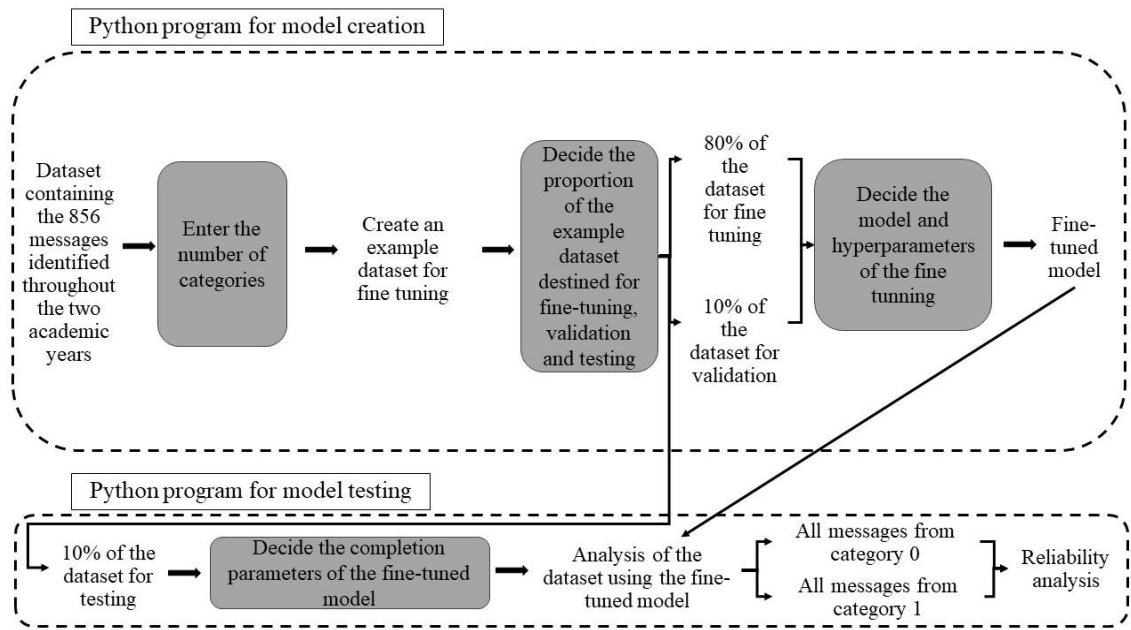

**Definition of sensitivity, specificity, precision, and F1 score**

Prior to defining these parameters, it is important to explain the following terms:

- True Positive (TP): The sentence is an engaging message (in identification), or the message belongs to a specific category (in classification) and is correctly labelled by the fine-tuned model.
- False Positive (FP): The sentence is not an engaging message, or the message doesn't belong to a specific category but is incorrectly labelled.
- True Negative (TN): The sentence is not an engaging message, or the message doesn't belong to a specific category and is correctly labelled.
- False Negative (FN): The sentence is an engaging message, or the message belongs to a specific category but is incorrectly labelled.

With this understanding, we can establish that:

- **Sensitivity (also known as Recall)**: Refers to the ability of the fine-tuned model to correctly classify that a sentence is indeed an engaging message, or that the message belongs to a certain category. Sensitivity is calculated as: TP/(TP + FN).
- **Specificity**: Denotes the model's capability to avoid false positives, that is, erroneously classifying a non-engaging sentence as engaging or misclassifying the category of a message. Specificity is defined as: TN/(TN + FP).
- **Precision:** Represents the proportion of correctly predicted positive cases (engaging messages or correct classifications) out of all predicted positive cases. It is calculated as: TP/(TP + FP).
- **F1 score:** It is the harmonic mean of precision and recall (or Sensitivity), providing a single performance metric that balances both. It is defined as: 2×(Precision×Recall)/(Precision+Recall).

**Supplemental Material References**